\documentclass[11pt]{article}
\usepackage[final]{acl}
\usepackage{graphicx}
\PassOptionsToPackage{options}{natbib}
\usepackage{amsmath}
\usepackage{mathptmx}
\usepackage{enumitem}
\usepackage{inconsolata}
\usepackage{todonotes}
\usepackage{multirow}
\usepackage{float}
\usepackage{array}
\usepackage{xcolor}
\usepackage{svg}
\usepackage[utf8]{inputenc}
\usepackage{array}
\usepackage{xcolor}
\usepackage[normalem]{ulem}
\useunder{\uline}{\ul}{}
\usepackage{polyglossia} 

\usepackage{tgtermes}
\usepackage[T1]{fontenc}
\newfontfamily{\writehi}[Script=Devanagari]{NotoSansDevanagari.ttf}
\setotherlanguage{odia}        
\setotherlanguage{gujarati}    
\newfontfamily\odiafont{NotoSansOriya.ttf}[Script=Odia]
\newfontfamily\gujaratifont{NotoSansGujarati.ttf}[Script=Gujarati]

\usepackage{longtable}   
\usepackage{booktabs}    
\usepackage{tabularray}
\newcommand{\benchmark}{\textsc{IndicParam}}
\newcommand{\cf}{\textit{c.f.~}}

\newcommand{\eg}{\emph{e.g.}}

\title{\benchmark: Benchmark to evaluate LLMs on low-resource Indic Languages}

\author{Ayush Maheshwari, Kaushal Sharma$^{\spadesuit}$, Vivek Patel$^{\spadesuit}$, Aditya Maheshwari$^{\spadesuit}$ \\
$^{\spadesuit}$Indian Institute of Management Indore, India ; $^{\spadesuit}$\textsc{BharatGen} \\
\texttt{ayush.hakmn@gmail.com, \{kaushals,vivekp,adityam\}@iimidr.ac.in}
}

\begin{document}
\maketitle
\begin{abstract}
While large language models excel on high-resource multilingual tasks, low- and extremely low-resource Indic languages remain severely under-evaluated. We present \benchmark, a human-curated benchmark of over 13,000 multiple-choice questions covering 11 such languages (Nepali, Gujarati, Marathi, Odia as low-resource; Dogri, Maithili, Rajasthani, Sanskrit, Bodo, Santali, Konkani as extremely low-resource). 
We evaluated 20 LLMs, both proprietary and open-weights, which reveals that even the top-performing \texttt{Gemini-2.5} reaches 58\% average accuracy, followed by  \texttt{GPT-5} (45) and \texttt{DeepSeek-3.2} (43.1). We additionally label each question as knowledge-oriented or purely linguistic to discriminate factual recall from grammatical proficiency. Further, we assess the ability of LLMs to handle diverse question formats-such as list-based matching, assertion-reason pairs, and sequence ordering-alongside conventional multiple-choice questions. \benchmark\ provides insights into limitations of cross-lingual transfer and establishes a challenging benchmark for Indic languages. 
The dataset is available at \url{https://huggingface.co/datasets/bharatgenai/IndicParam}\footnote{Scripts to run benchmark are present at \url{https://github.com/ayushbits/IndicParam}}.

\end{abstract}

\section{Introduction}

\begin{figure*}[!ht]
    \centering
    \includegraphics[width=0.9\linewidth]{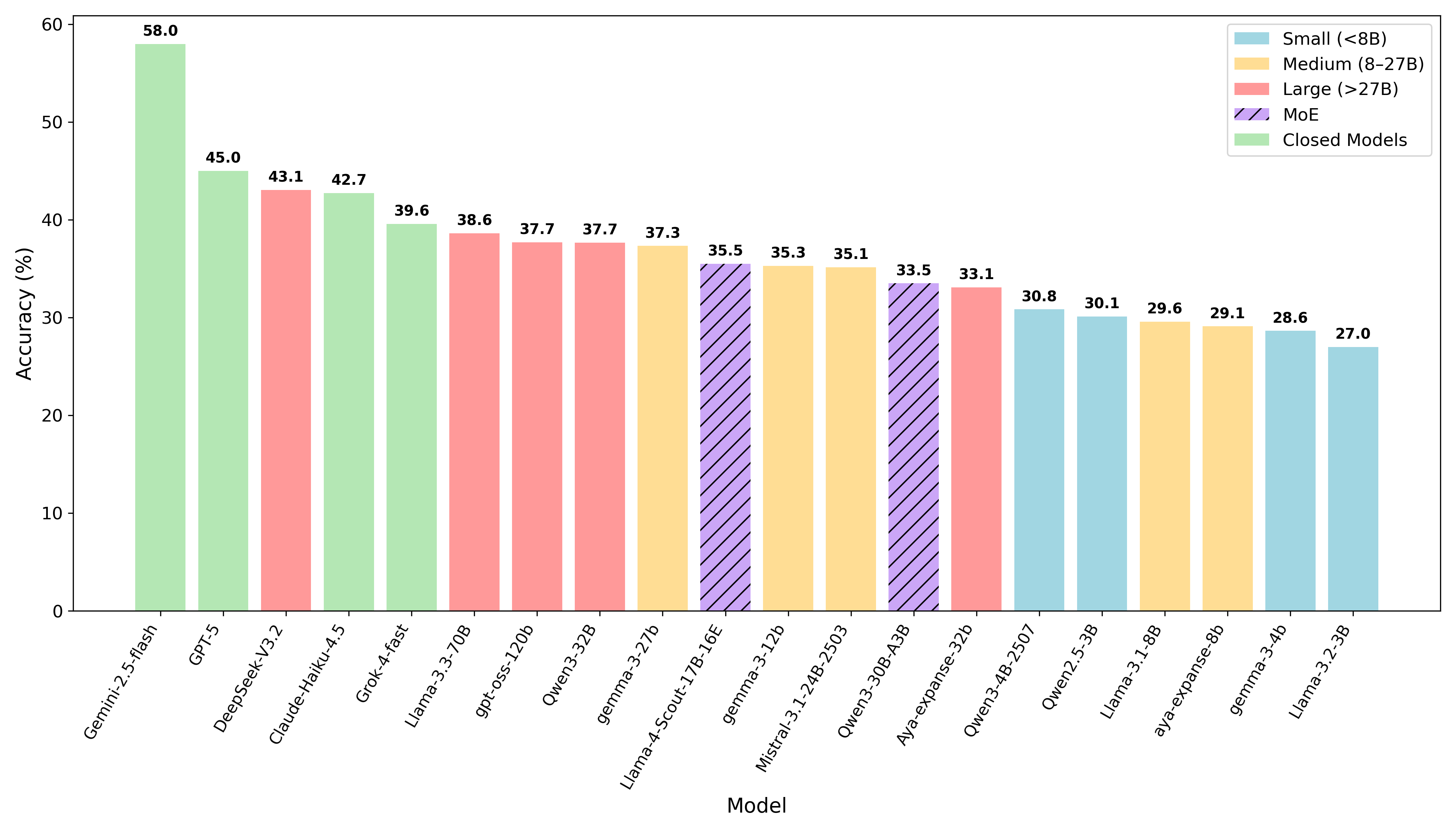}
    \caption{Average performance for all languages on \benchmark. Except DeepSeek-V3.2, all closed models outperform best performing open-weight models. }
    \label{fig:all-model-accuracy}
\end{figure*}
Large language models (LLMs) have delivered state-of-the-art results across a range of multilingual tasks, particularly in high- and medium-resource settings such as translation, named entity recognition, and question answering \cite{liu2024omgeval}. However, systematic evaluation for low and extremely-low-resource Indic languages remains limited. Existing benchmarks for Indic languages have significant coverage for major languages (\eg\ Hindi, Tamil, Telugu)~\cite{singh2024indicgenbench, kakwani2020indicnlpsuite}, yet they provide only partial or no support for several understudied languages. Additionally, they rarely include  coverage for code-mixed usage and evaluation.


\begin{table}[!]
\centering
\resizebox{0.45\textwidth}{!}{
\begin{tabular}{@{}lcc@{}}
\toprule
\textbf{Language} & \textbf{\#Words (in M)} & \textbf{\#Speakers (in M)} \\ \midrule
Nepali            & 1642.9                         & 19.4                              \\
Marathi           & 1541.2                         & 99.1                              \\
Gujarati          & 934.1                          & 60.3                              \\
Odia              & 333.8                          & 42.6                              \\
Sanskrit          & 16.9                           & 3.1                               \\
Maithili          & 14.0                           & 14.3                              \\
Konkani           & 2.8                            & 2.6                               \\
Santali           & 0.8                            & 7.7                               \\
Bodo              & 0.8                            & 1.5                               \\
Dogri             & 0.6                            & 2.8                               \\
Rajasthani        & -                              & 25.8  \\ \midrule
Total &       4480                                  & 279.1 \\ \bottomrule 
\end{tabular}
}
\caption{Number of words (in millions) present in FineWeb2 corpus and number of speakers (in millions) for each language considered in \benchmark.}
\label{tab:words-speakers}
\end{table}

In Table \ref{tab:words-speakers}, we present word counts for the \benchmark\ languages within FineWeb2 \cite{penedo2025fineweb2pipelinescale}. FineWeb2 is a cleaned pretraining corpus comprising of approximately 3 trillion words from around 1800 languages (excluding English) gathered from more than 100 Common Crawl snapshots. For the 11 languages analyzed in  this benchmark, the FineWeb2 corpus contains 4.49 billion words. Four languages-Nepali, Marathi, Gujarati, and Odia-account for 4.45 billion of these words, leaving only 36 million words for the remaining six languages. Collectively, these 11 languages represent about 280 million speakers across the Indian subcontinent, yet their aggregate web presence is disproportionately small relative to their speaker base. Based on the availability of web corpus, we classify Marathi, Gujarati, Nepali and Odia as low-resource languages while remaining seven languages as extremely low-resource languages.

\begin{table}[h!]
\centering
\begin{tabular}{|p{0.96\linewidth}|}
\hline
\textbf{Low resource (4)}:  Nepali, Gujarati, Marathi, Odia \\[4pt]
\textbf{Extremely-low resource (7)}: Dogri, Maithili, Rajasthani, Sanskrit, Bodo, Santali, Konkani\\
\hline
\end{tabular}
\end{table}
India, with a population of approximately 1.4 billion, is home to over 120 languages, at least 30 of which have one million or more speakers \cite{census2018language}. These languages span multiple families, for instance, Marathi and Hindi belong to the Indo-European branch; Kannada and Telugu are Dravidian; Santali is Austro-Asiatic; and Bodo is Sino-Tibetan. The Indian Constitution recognizes 22 languages in the Eighth Schedule, yet most existing benchmarks concentrate on 11 major languages that together account for more than 93\% of speakers, largely reflecting the limited availability of web-scale data for others \cite{IAMAI2024}. With recent advances in multilingual LLMs, language understanding has improved for many of these widely used languages \cite{hu2020xtreme,Qin2025}. However, despite a few benchmarks covering the majority set of 11 languages, there exists almost no evaluation resources for other Indic languages \cite{kumar2022indicnlg, madasu2023mukhyansh}.



To address these limitations, this paper introduces \benchmark, a human-supervised evaluation set comprising more than 13,000 questions for 11 low- and extremely low-resource Indic languages. We also include a code-mixed Sanskrit-English language set evaluation.
We hypothesize that despite limited pretraining data in several of these languages, LLMs can exploit cross-lingual transfer among typologically and script-wise related languages to improve performance on these target languages. To separate language understanding from factual recall, each question is labeled as either linguistic (\eg, morphology, syntax, semantics, discourse) or knowledge-based (world facts and entities).

We evaluate 20 LLMs spanning open- and closed-source systems and a range of parameter scales and architectures on \benchmark. The evaluated models includes closed models such as GPT-5, Gemini-2.5, Claude-Haiku-4.5, and open weight models such as Llama4-Scout, DeepSeek-V3.2 among others. In Figure \ref{fig:all-model-accuracy}, we present the average performance of all models. Larger models with substantial exposure to multilingual Indic corpora tend to outperform smaller models. However, no model exceeds a 58\% average score on \benchmark.
Gemini-2.5 achieves the highest overall accuracy at 58, followed by GPT-5 at 45, Claude-Haiku at 42.7 and Grok-4 at 39.6, indicating that model robustness on these languages remains limited even among state-of-the-art proprietary models. 
Open-weight models narrows the gap with DeepSeek-V3.2 \cite{deepseekai2024deepseekv32}, which is a 685B parameter model (43.1\%) while Llama 3.3-70B attains average accuracy of 39.6. 
Overall, results indicate that multilingual performance for Indic languages remains challenging across both closed and open models, with substantial room for improvement in cross-lingual generalization.
Our contributions can be summarised as follows:

\noindent 1. We present \benchmark, a benchmark consisting of 13K+ questions on 11 low- and extremely-low resource Indic languages, as well as a code-mixed English-Sanskrit variant.

\noindent 2. We provide granular annotations for each instance in the benchmark, including question-type and question-category labels, to enable systematic analysis of LLM capabilities.

\noindent 3. We conduct an extensive evaluation encompassing 20 LLMs, including both proprietary closed-source and open-weight models.

\section{Related Work}
LLMs have demonstrated substantial improvements in language understanding performance for several Indic languages \cite{sarvamSarvamSovereign, pundalik2025param1bharatgen29bmodel}. Multiple Indic multilingual benchmarks evaluate such progress, primarily targeting major mid- and high-resource languages \cite{singh2025indic, singh2024indicgenbench, maheshwari-etal-2024-samayik}. IndicGenBench \cite{singh2024indicgenbench} proposes a benchmark on 29 Indic languages, including those examined in this study. Its evaluation suite includes tasks such as reading comprehension and translation, which are primarily adapted from English benchmark datasets like XQUAD \cite{artetxe2020cross} and FLORES \cite{goyal2022flores}. IndicGenBench QA tasks restrict answers to spans supported by the given passage. In contrast, \benchmark\ extends this evaluation paradigm by assessing both language understanding and domain-specific general knowledge of LLMs using questions drawn from graduate-level examinations.

IndicGLUE \cite{kakwani2020indicnlpsuite} is a suite of supervised NLP tasks for prominent Indic languages, with a particular emphasis on Hindi, including classification tasks such as news categorization and headline prediction, as well as natural language inference, among others. IndicXTREME \cite{doddapaneni2023towards} broadens task coverage to named entity recognition, question answering, and paraphrase detection, and expands language coverage by translating and curating benchmarks in 11 Indic languages; however, many long-tail languages remain outside its scope. Recent Indic QA benchmarks further diversify evaluation: ParamBench \cite{maheshwari2025parambench} focuses on graduate-level questionnaires with varied question types in Hindi, while MILU \cite{verma2025milu} contains 79K questions across 11 Indian languages, including Odia, Gujarati, and Marathi, which are also covered in \benchmark. Sanskriti \cite{maji2025sanskriti} and Pariksha \cite{watts2024pariksha} are English-language benchmarks designed to assess the socio-cultural alignment of LLMs. Drishtikon \cite{maji2025drishtikon} further extends this line of work to multimodal cultural understanding across 15 Indic languages.

Despite this progress, most benchmarks concentrate on the majority of 11 widely used Indian languages, with very limited coverage for extremely low-resource languages. In this paper, we introduce graduate-level language understanding questionnaires in 4 low-resource and 7 extremely low-resource languages, enabling targeted assessment of both language competence and general knowledge for underrepresented Indic languages.

\section{\benchmark}
\benchmark\ is a high-quality QA benchmark consisting of 13,207 questions that evaluates language understanding and general knowledge of multilingual LLMs in 11 Indic languages. 
These languages are spoken by approximately 280 million people across the Indian subcontinent, of whom 10 languages are recognized as scheduled languages in the Indian Constitution. Rajasthani, though not a scheduled language, is spoken by roughly 25 million people. We also include a QA set of code-mixed version of Sanskrit where questions include a mix of English and Sanskrit.

\begin{table}[]
\resizebox{.5\textwidth}{!}{%
\begin{tabular}{@{}ccccc@{}}
\toprule
\textbf{Language} & \textbf{\#Ques} & \textbf{LU(\%)} & \textbf{Script}      & \textbf{Code}        \\ \midrule
Nepali            & 1038                 &     18.8         & Deva           & npi                  \\
Marathi          &   1245                    & 4.7                  & Deva           & mar                  \\
Gujarati          &   1044                    &   0.6               & Gujarati             & guj                  \\
Odia              & 577                      &     20.3             & Orya                 & ory                  \\ \midrule
Maithili          & 1286                 &             10.1     & Deva           & mai                  \\
Konkani           &   1328                   &             2.5     & Deva           & gom                  \\
Santali           &       873               &     11.3             & Olck                 & sat                  \\
Bodo              &          1313            &    -              & Deva           & brx                  \\
Dogri             & 1027                 &     18.3             & Deva           & doi                  \\
Rajasthani        & 1190                 &            27.8      & Deva           & -                    \\
Sanskrit          & 1315                 &            20.7      & Deva           & san                  \\
Sans-Eng          & 971                  &             11.5     & -                    & -                    \\ \midrule
\textbf{Total}             & \textbf{13207}  &                  & \multicolumn{1}{l}{} & \multicolumn{1}{l}{} \\ \bottomrule
\end{tabular}
}
\caption{Distribution of question-answer pairs for different languages in \benchmark. `Sans-Eng' denotes a separate set of Sanskrit-English code-mixed question-answer pairs. LU refers to \% of manually classified language-understanding questions; rest are categorized as knowledge-related (\cf Section \ref{sec:quesclassification}).} 
\label{tab:stats}
\end{table}
The language-wise distribution of questions is shown in Table \ref{tab:stats}. The questions are collected from UGC-NET\footnote{\url{https://ugcnet.nta.ac.in}}, which is a nationwide examination administered by a government agency to determine eligibility for PhD admission and appointment to teaching positions in Indian universities and colleges. The exam is offered in around 85 subjects and is conducted twice annually. Each test consists of two papers composed of multiple-choice questions (MCQs). 
We developed the dataset by downloading question papers and answer keys from the official examination website for the period between 2012-2018.

Each language comprises multiple PDF question papers, many machine-readable and a subset with non-selectable text. Document layouts vary, with some single-column and most two-column. To ensure uniform accessibility, all PDFs were processed with a proprietary OCR system, and the extracted text was used for downstream curation and annotation. To the best of our knowledge, this corpus has not appeared in prior LLM benchmarking studies and constitutes a newly curated, human-authored dataset explicitly designed for graduate-level evaluation in Indic contexts. We describe annotation setup and team structure in Section \ref{sec:team} in Appendix.

\subsection{Question Classification}\label{sec:quesclassification}

In addition to annotating question-answer pairs in \benchmark, each question is assigned to one of two categories: (a) \textbf{language understanding} (LU) that includes questions related to linguistics and grammar, and (b) \textbf{general knowledge} (GK) targeting fact-based queries. 
Given the low- and extremely low-resource nature of the studied languages, this evaluation investigates model performance under data scarcity conditions and examines the extent to which cross-lingual transfer mechanisms can compensate for limited in-language training data. 

The distribution of questions across these categories for each language is presented in Table \ref{tab:stats}. The LU column reports, for each language, the proportion of questions that test language understanding while the remaining questions belonging to the GK category and focusing on factual knowledge. The share of LU questions varies substantially, from under 1\% in Gujarati to nearly 28\% in Rajasthani, with several other languages such as Odia and Sanskrit also exhibiting over 20\% LU coverage. In Table \ref{tab:examplequestions}, we present example questions from different languages and corresponding question class and type.

\subsection{Question Type Classification}
\label{Qtypeclass}
We annotate each question-answer pair with its corresponding question type by classifying them into six primary categories that capture the diversity of assessment formats (see Table \ref{tab:stats-qtypes}):  multiple-choice questions (MCQ), assertion and reasoning (A\&R), list matching, blank filling, identify incorrect statement (IS), and ordering. The majority (73\%) of the questions are MCQs followed by list matching (9\%), ordering (6.5\%), A\&R (6.1\%), IS (4.1\%) and blank filling (1.1\%). The language wise breakdown of all languages remains the same except for Bodo, Gujarati,  and Dogri, where the proportion of MCQs remain 36.6\%, 33\% and 21.7\%, respectively, which are significantly lower than other languages in the benchmark. The detailed statistics of language-wise splits of question types is present in Table \ref{tab-app:stats-qtypes}.
In Section \ref{app:sec-qtype}, we present detailed discussion on question type-wise results for different models.

\begin{table}[]
\centering
\resizebox{.4\textwidth}{!}{%
\begin{tabular}{@{}lc@{}}
\toprule
\textbf{Question Type}       & \textbf{\#Questions} \\ \midrule
Multiple-choice              & 9653                 \\
Assertion \& Reason          & 811                  \\
List Matching                & 1185                 \\
Fill in the blanks           & 157                  \\
Identify incorrect statement & 545                  \\
Ordering                     & 856                  \\ \midrule
\textbf{Total}               & \textbf{13207}       \\ \bottomrule
\end{tabular}
}
\caption{Distribution of questions across all languages in \benchmark.}
\label{tab:stats-qtypes}
\end{table}

\subsection{Quality control and Verification}
We used a multistep verification pipeline for linguistic precision and integrity of data. Primary data was taken from high-resolution, native PDFs from the official UGC NET portal to minimize extraction noise. Text extraction was script-tailored: LLMWhisperer\footnote{\url{https://unstract.com/llmwhisperer/}} for Devanagari scripts-Sanskrit, Maithili, Marathi-to preserve complex conjuncts, and Surya OCR \cite{paruchuri2025surya} for Gujarati and Odia, followed by intensive manual correction by experts.

A post-verification audit on a random 5\% sample (10,427 words and 61,547 characters) reported a Word Error Rate (WER) of 0.68\% and a Character Error Rate (CER) of 0.12\%, confirming high reliability of the processed text.  Please refer to the Appendix section \ref{subsec: Quality Control and Verification} for more details.



\noindent

\subsection{Setup}

We evaluated 20 state-of-the-art models spanning both open-weight and proprietary LLMs. The open-weight models range from 3 billion to over 685 billion parameters, including both dense and mixture-of-experts (MoE) architectures. These models collectively represent the state of the art in multilingual reasoning, code understanding, and multimodal generation. All models were used in their instruction-tuned configurations and evaluated under a zero-shot prompting setup. The prompt used during evaluation is present in Table \ref{tab:zero-shot-prompting}.
We apply a single, standardized prompt template across all languages and models. We also experimented with alternative prompt formulations, including prompts written in different languages; however, these variations did not meaningfully affect the results

Open-weights models are retrieved from publicly available checkpoints at HuggingFace and use transformers library for inference. We load parameter weights in bf16 precision for all open-weight models except Llama4-Scout which was loaded with 8-bit quantization. We use greedy decoding while generating predictions by setting max-tokens as 50, temperature as 0, do\_sample flag to false and batch size is set to 16. We disable thinking mode for Qwen3-MoE and keep reasoning levels at low for GPT-OSS-120B. For Gemma and Mistral, we follow their respective strategies from response generation instead of transformers pipeline function. We use OpenRouter (\url{https://openrouter.ai})  to generate predictions for closed models.
We next describe the models included in our evaluation.

\noindent 1.  \textbf{Qwen Series~\cite{yang2025qwen3technicalreport}:} We evaluated multiple models of the Qwen family, including \texttt{Qwen2.5-3B}, \texttt{Qwen3-4B-2507}, \texttt{Qwen3-30B-A3B}, and \texttt{Qwen3-32B}. The Qwen 3 series integrates substantial architectural and pre-training improvements over the 2.5 generation, trained on roughly 36 trillion multilingual tokens spanning 119 languages. The \texttt{Qwen3-30B-A3B} variant adopts a mixture-of-experts (MoE) design with 3 billion active parameters per forward pass, providing strong efficiency while maintaining performance parity with larger dense models.

\noindent 2. \textbf{Llama Series~\cite{grattafiori2024llama3herdmodels}:} We evaluated Meta’s \texttt{Llama-3.2-3B}, \texttt{Llama-3.1-8B}, and \texttt{Llama-3.3-70B}. These models are trained on a mixture of high-quality multilingual and code data totaling approximately 15 trillion tokens. We also evaluated on  \texttt{Llama-4-Scout-17B-16E} which is an MoE model having a 17 billion active parameters with 16 experts with a total of 109 billion model parameters. This model is trained  on 40T tokens specifically designed to improve multi-lingual text and images.

\noindent 3. \textbf{Gemma Series~\cite{gemmateam2025gemma3technicalreport}:} We evaluated \texttt{Gemma-3} models at 4B, 12B, and 27B parameter scales. Developed by Google DeepMind, the Gemma 3 family employs hybrid instruction-fine-tuning strategies and trained on diverse multilingual datasets. These models have demonstrated strong alignment and competitive performance across multilingual and reasoning benchmarks.

\noindent 4. \textbf{Aya Expanse~\cite{aryabumi2024aya23openweight}:} The \texttt{Aya-Expanse-8B} and \texttt{Aya-Expanse-32B} models are instruction-tuned multilingual LLMs. These models prioritize cultural and linguistic diversity and are trained on extensive parallel datasets that include over 100 different languages.

\noindent 5. \textbf{Mistral 3.1~\cite{MistralAI}:} We evaluated the \texttt{Mistral-Small-3.1-24B-2503} which is a 24 billion parameter multi-lingual model achieving competitive performance while maintaining fast inference throughput.

\noindent 6. \textbf{DeepSeek V3.2~\cite{deepseekai2024deepseekv32EXP}:} We evaluated \texttt{DeepSeek-V3.2-Experimental} model consisting of 685B parameters. It employs specialized routing and sparsity mechanisms to reduce inference costs while preserving high-quality reasoning, code synthesis, and mathematical reasoning performance.

\noindent 7. \textbf{GPT-OSS-120B~\cite{openai2025gptoss120bgptoss20bmodel}:} We examined the \texttt{GPT-OSS-120B} model, an open-weight MoE model by OpenAI, trained on trillions of tokens with a focus on STEM, coding, and general knowledge. It has a total of 116.8B parameters and 5.1B active parameters per token per forward pass.

\noindent 8. \textbf{Frontier Proprietary Models:} For completeness, we also benchmarked several leading closed-source systems accessible through public APIs: \texttt{GPT-5}, \texttt{Claude-Haiku-4.5}, and \texttt{Grok-4-Fast}. 
\\
\noindent -- \textit{GPT-5} (OpenAI) represents the current flagship frontier model, incorporating multimodal reasoning and tool-use capabilities. 
\\
\noindent -- \textit{Claude-Haiku-4.5} (Anthropic) is a lightweight yet high-accuracy variant of the Claude 4.5 family designed for efficiency-critical environments. 
\\
\noindent -- \textit{Grok-4-Fast} (xAI) is optimized for rapid inference and contextual awareness within streaming conversational environments.
\\
\noindent -- \textit{Gemini-2.5-Flash~ \cite{comanici2025gemini}} (Google) is a sparse MoE model and a distilled version of a larger Gemini-2.5 model.

\begin{table*}[]
\centering
\resizebox{\textwidth}{!}{
\begin{tabular}{@{}clccccccccc@{}}
\toprule
\textbf{Size}                & \multicolumn{1}{c}{\textbf{Model}} & \textbf{Dogri}         & \textbf{Maithili}      & \textbf{Rajasthani}    & \textbf{Sanskrit}      & \textbf{Sans-Eng}   & \textbf{Bodo}       & \textbf{Santali}    & \textbf{Konkani}    & \textbf{Average}       \\ \midrule
\multirow{4}{*}{Small}  & Qwen2.5-3B                         & 32.1                   & 28.8                   & 28.6                   & 31.9                   & {\ul \textit{34.8}} & {\ul \textit{28.3}} & 33.1                & {\ul \textit{31.4}} & 30.9                   \\
                             & Llama-3.2-3B                       & 27.9                   & 21.9                   & 26.3                   & 28                     & 27.9                & 27.8                & 27.4                & 28.8                & 26.9                   \\
                             & Gemma-3-4b                         & 30.5                   & 24.5                   & 27.6                   & 29.1                   & 32.2                & 25.8                & 31.7                & 29.1                & 28.5                   \\
                             & Qwen3-4B                           & {\ul \textit{34.6}}    & {\ul \textit{31.6}}    & {\ul \textit{30}}            & {\ul \textit{33.8}}    & 33.6                & 25.1                & {\ul \textit{33.8}} & 29.6                & {\ul \textit{31.2}}    \\ \midrule
\multirow{4}{*}{Medium} & Aya--8b                            & 31.5                   & 29.9                   & 30.8                   & 29.4                   & 29.9                & 28.1                & 30.8                & 26.9                & 29.5                   \\
                             & Llama-3.1-8B                       & 26.4                   & 29                     & 31.8                   & 32.2                   & 34                  & 26.8                & 30.6                & 29.5                & 30.0                   \\
                             & Gemma-3-12b                        & 37.2                   & 31.6                   & 35                     & 37.1                   & 41.8                & {\ul \textit{30.6}} & 33.7                & 35.7                & 35.1                   \\
                             & Mistral3.1-24b                     & {\ul \textit{38}}      & {\ul \textit{37.4}}    & 34.8                   & {\ul \textit{40}}      & 44.1                & 30.1                & 35.7                & 32                  & 36.2                   \\
                             & Gemma-3-27b                        & 37.6                   & 34.4                   & {\ul \textit{39.8}}    & 38.6                   & {\ul \textit{49.1}} & 28.9                & {\ul \textit{36.7}} & {\ul \textit{36.5}} & {\ul \textit{37.3}}    \\ \midrule
\multirow{4}{*}{Large}  & Aya-32b                            & 35.2                   & 33.2                   & 34.6                   & 35.2                   & 39                  & 27.2                & 35.6                & 33.1                & 33.7                   \\
                             & Qwen3-32B                          & {\ul \textit{43.4}}    & 34                     & 33.8                   & 39.2                   & 48.6                & {\ul \textit{31.8}} & 39.7                & 34.5                & 33.9                   \\
                             & Llama-3.3-70B                      & 37.3                   & {\ul {37.5}}    & {\ul \textit{38.2}}    & {41.7}    & {51.6} & 28.6                & {40.1} & {37.3} & {37.6}    \\
                             & DeepSeek-3.2                           & 43.3                   & {\ul \textit{41.4}}                   & 40.1                   & {\ul 51.3}             & {\ul 61.4}          & \textbf{\textit{36.4}}       & \textbf{\textit{41.5}}       & {\ul \textit{37.6}}                & {\ul 43.7}  \\ \midrule
\multirow{3}{*}{MoE}         & Qwen3-A3B                          & 35.7                   & 30.9                   & 31.6                   & 36.7                   & 44.2                & {\ul \textit{29.9}} & 35.9                & 28.3                & {\ul \textit{38.6}}          \\
                             & Llama-4-Scout                      & 30.5                   & 31.3                   & 34.8                   & {\ul \textit{44}}      & {\ul \textit{48.7}} & 25.9                & {\ul \textit{36.3}} & 34.6                & 35.4                   \\
                             & gpt-oss-120b                       & {\ul \textit{39.7}}    & {\ul \textit{35.6}}    & {\ul \textit{39.4}}    & 40.2                   & 48.1                & 28.3                & 35.6                & {\ul \textit{35.4}} & 37.4                   \\ \midrule
\multirow{4}{*}{Closed}      & GPT-5                              & {\ul 45.7} & {\ul 43.4} & {\ul 44.6} & {\ul 54.8} & {\ul 64.6}       & {\ul 31.6}          & 40.4                & { 39}            & {\ul 45.1} \\
                             & Grok-4-fast                        & 40.2                   & 38.1                   & 39.1                   & 42.9                   & 54.2                & 30.6                & 38.8                & 37.4                & 39.7                   \\
                             & Claude-4.5                         &  44.8             & {42.4}             & { 44.1}             & 47.8                   & 58.1                & 28.1                & {\ul 40.4}          & {\ul 42.1}       & 43.0                   \\
                             
& Gemini-2.5 & \textit{\textbf{54.2}} & \textit{\textbf{54}} & \textit{\textbf{66.6}} & \textit{\textbf{72.4}} & \textit{\textbf{80.3}} & \textit{\textbf{48.1}} & \textit{\textbf{52.9}} & \textit{\textbf{48.4}}    & \textit{\textbf{59.2 }} \\
                              \bottomrule
\end{tabular}
}
\caption{Performance on extremely-low resource languages on \benchmark. We \textbf{bold} and \textit{italicize} the best overall performance, {\ul underline} and \textit{italicize} the best performance in each model-size category and {\ul underline} the second best overall performance. }
\label{tab:perf-extremely-low}
\end{table*}
\section{Results and discussion}
\subsection{Overall Model Performance}



The evaluation results presented in Figure~\ref{fig:all-model-accuracy} demonstrate that average performance on \benchmark\ for 20 evaluated models remains moderate, though several interesting patterns emerge when grouped by model scale. Among the small-capacity models (<8B parameters), the Qwen family performs best: \texttt{Qwen3-4B} and \texttt{Qwen2.5-3B}  achieves the  accuracy of 30.8 and 30.1 respectively. Other models in this category, including \texttt{Gemma-3-4B}, and \texttt{Llama-3.2-3B},  remain around the 27-30\% range. 

In the medium-size category (8B-27B), performance improves considerably, with \texttt{Gemma-3-27B} leading at 37.3\%, followed closely by \texttt{Gemma-3-12B} (35.3) and  \texttt{Mistral-Small-3.1-24B} (35.1). Other models like \texttt{Aya-Expanse-8B} and \texttt{Llama-3.1-8B} score less than 30\%. These findings underscore that model scale beyond 8B yields measurable gains, but pre-training coverage and multilingual representation continue to play a decisive role in performance.


Among the larger and frontier models ($\geq$27B), overall accuracy rises substantially. The best performance is achieved by \texttt{Gemini-2.5}, which attains an average of 58\%, followed by \texttt{GPT-5} (45), \texttt{DeepSeek-V3.2} (43.1), \texttt{Claude-Haiku-4.5} (42.7), and \texttt{Grok-4-Fast} (39.6). Open-weight models such as \texttt{Llama-3.3-70B} and \texttt{GPT-OSS-120B} achieve 38.6 and 37.7, respectively, while the \texttt{Llama-4-Scout-17B-16E} models remain slightly lower at 35.5. These results indicate that while data and parameter size enhance overall accuracy, the highest-performing models also benefit from refined fine-tuned on multilingual or Indic-rich corpora.

Overall, the evaluation reveals a clear upward trajectory in accuracy with increasing model scale, yet even the most capable frontier systems such as \texttt{Gemini-2.5}, \texttt{GPT-5} and \texttt{DeepSeek-V3.2} demonstrate that substantial headroom remains for further enhancement in Indic language understanding. The results affirm that cross-lingual generalization, fine-tuning quality, and the underlying token distribution remain central determinants of downstream accuracy.

\begin{table*}[!h]
\centering
\resizebox{0.7\textwidth}{!}{
\begin{tabular}{@{}clccccc@{}}
\toprule
\textbf{Size}                & \multicolumn{1}{c}{\textbf{Model}} & \textbf{Nepali}     & \textbf{Gujarati}   & \textbf{Marathi}    & \textbf{Odia}       & \multicolumn{1}{l}{\textbf{Average}} \\ \midrule
\multirow{4}{*}{Small}  & Qwen2.5-3B                         & 27.1                & {\ul \textit{29.5}} & 27.9                & 28.8                & 28.2                                 \\
                             & Llama3.2-3B                       & 26.2                & 26.1                & 28.6                & 27                  & 27.1                                 \\
                             & Gemma-3-4b                         & {\ul \textit{29}}   & 26.2                & 29.9                & 31.5                & 28.9                                 \\
                             & Qwen3-4B                           & 28.7                & 27.4                & {\ul 30.5}          & {\ul \textit{34.7}} & {\ul \textit{29.8}}                  \\ \midrule
\multirow{4}{*}{Medium} & Aya-8b                            & 28.2                & 27.7                & 28.3                & 28.1                & 28.1                                 \\
                             & Llama3.1-8B                       & 25.5                & 27.2                & 32.2                & 29.3                & 28.7                                 \\
                             & Gemma-3-12b                        & 34.8                & {\ul \textit{33.8}} & 38.4                & 34.5                & 35.6                                 \\
                             & Mistral3.1-24b                     & 34.9                & 32.6                & 32.9                & 27.6                & 32.6                                 \\
                             & Gemma3-27b                        & {\ul \textit{39.3}} & 31.1                & {\ul \textit{39.9}} & {\ul \textit{39.9}} & {\ul \textit{37.4}}                  \\ \midrule
\multirow{4}{*}{Large}  & Aya-32b                            & 31.6                & 30.7                & 35.5                & 34.5                & 33.0                                 \\
                             & Qwen3-32B                          & 33                  & 28.4                & 33.2                & 28.9                & 31.2                                 \\
                             & Llama3.3-70B                      &35   & 35.9 & 40.8 & 40.7 & 37.9                  \\
                              & DeepSeek-3.2                           & {\ul \textit {39.7}}                & {\ul \textit{ 39.5}}          & {\ul \textit{ 44.6}}          & {\ul \textit {42.1}}                & {\ul \textit {41.6}}                                 \\
                             \midrule
\multirow{3}{*}{MoE}         & Qwen3-A3B                          & {\ul \textit{37.2}} & 33.5                & 43.5                & {\ul \textit{40.4}} & {\ul \textit{38.7}}                  \\
                             & Llama-4-Scout                      & 34.6                & 27.3                & {\ul \textit{43.5}} & \textit{35}         & 35.5                                 \\
                             & gpt-oss-120b                       & 37.1                & {\ul \textit{37}}   & 39.7                & 39.9                & 38.3                                 \\ \midrule
\multirow{4}{*}{Closed}     
                             & GPT-5                              & {43.4}          & {\ul 40}         & {\ul 48.4}       & {\ul 48.7}       & {\ul 44.9}                        \\
                             & Grok-4-fast                        & 40.9                & 35.9                & 39.8                & 40.7                & 39.2                                 \\
                             & Claude-4.5                         & {\ul 43.8}       & 37                  & 43.9                & { 44}            & { 42.0}                           \\ 
& Gemini-2.5 &\textbf{\textit{52}}	& \textbf{\textit{51.7}} & 	\textbf{\textit{55.7}} & 	\textbf{\textit{64.5}}   & \textbf{\textit{54.9}}  \\               

\bottomrule
\end{tabular}
}
\caption{Performance on low resource languages on \benchmark. We \textbf{bold} and \textit{italicize} the best overall performance, {\ul underline} and \textit{italicize} the best performance in each model-size category and {\ul underline} the second best overall performance. }
\label{tab:perf-low}
\end{table*}

\subsection{Performance on Low- and Extremely Low-Resource Indic Languages}

Table~\ref{tab:perf-extremely-low} and Table~\ref{tab:perf-low} compare contemporary LLMs on eight extremely low-resource Indic languages (Dogri, Maithili, Rajasthani, Sanskrit, Sanskrit--English code-mixed, Bodo, Santali, and Konkani) and four higher (yet still low-resource) Indic languages (Nepali, Gujarati, Marathi, and Odia) respectively. 
On extremely low-resource languages, \texttt{Gemini-2.5} achieves the highest average of 59.2\%, followed by \texttt{GPT-5} (45.1), open-weight \texttt{DeepSeek-3.2} at 43.7. No model exceeds 53 on the most underrepresented languages (Bodo and Santali), with most models remaining below 36\%. \texttt{Gemini-2.5} shows strong performance on Sanskrit (72.4) and Sanskrit--English code-mixed text (80.3). Despite Gemini-2.5 performing remarkably better than other LLMs, the remaining models exhibit substantially lower performance, suggesting that strong Indic-language capability is not yet widespread across current LLMs.

With the exception of Rajasthani and Sanskrit, all models including \texttt{Gemini} scores less than 54\% for all other languages. Notably, even very large parameter Deepseek model scores under 44\%, aside from the Sanskrit variants. 
The relatively stronger results for Rajasthani may stem from its linguistic proximity to Hindi, which is typically well represented in model pretraining corpora. Despite limited web-scale data availability, Sanskrit is a well-studied language in the NLP literature, which may partially explain its comparatively better performance. 


Among smaller models, the Qwen family outperforms Gemma and Llama models. In contrast, \texttt{Gemma3-27B} surpasses \texttt{Qwen3-32B}, which is counter-intuitive to the assumption that Qwen3's larger pretraining corpus necessarily yields consistent downstream gains. Overall, within the open-weight models, \texttt{DeepSeek} performs best, suggesting benefits from its architectural choices and larger parameter scale.

In the low-resource group, \texttt{Gemini-2.5} again leads with 54.9\%, ahead of \texttt{GPT-5} (44.9), \texttt{Claude-4.5} (42.0) and \texttt{DeepSeek-3.2} (41.6). Among other open models, \texttt{Gemma-3-27b} (37.4) and the MoE \texttt{Qwen3-A3B} (38.7) perform best in their respective size categories. Notably, \texttt{Gemini-2.5} achieves the highest performance in Odia (64.5), followed by Marathi (55.7) having a 8.8 points drop. In contrast, other models exhibit a much smaller performance gap between these two languages.

Except for \texttt{Gemini}, all other models score below 50\% across each of the four languages considered. Small-size models achieve their highest score in Odia (34.7), and medium-size models likewise obtain their best result in Odia (39.9). Among larger open-weight models, \texttt{DeepSeek} attains the strongest performance, scoring 44.6 on Marathi. Overall, these findings suggest that improving performance on low-resource Indic languages likely requires both stronger general pretraining data and targeted, language-specific resources.



\subsection{Performance on GK and LU Tasks} 

We distinguish between two complementary capabilities: (i) \textit{General Knowledge} questions that require factual reasoning and culturally grounded world knowledge (Table~\ref{tab:qclass-gk}), and (ii) \textit{Language-Specific Understanding} questions that probe grammatical, lexical, morphosyntactic, and discourse-level proficiency within each Indic language (Table~\ref{tab:qclass-LU}). LU questions forms a very small percentage of overall questions, particularly for Gujarati (0.6\%).

\subsubsection{General Knowledge}
\label{sec:gk-performance}
Frontier models exhibit clear separation from the rest of the models. \texttt{Gemini-2.5} achieves the highest average (58), followed by \texttt{GPT-5} (54.8), \texttt{Claude-4.5} (43.5) and \texttt{DeepSeek-3.2} (43.3). Among open-weight systems, \texttt{Llama-3.3-70B} (39.3) and \texttt{gpt-oss-120b} (38) perform competitively, while \texttt{Grok-4-fast} attains 40.3. Large gains are observed with scale: medium-sized models (8B--27B) top out at 37.9 (\texttt{Gemma-3-27b}), and small models (<8B) remain below 32 (\texttt{Qwen3-4B}: 31.3). As shown in Table~\ref{tab:qclass-gk}, \texttt{Gemini} dominates Sanskrit (74.4) and Sanskrit--English code-mixed (80.3) contexts, reflecting superior classical-language representation.

\subsubsection{Language-Specific Understanding}
\label{sec:lang-performance}
Performance on pure linguistic tasks is higher than GK questions on all languages except Sanskrit. Except \texttt{Gemini}, none of the models rarely exceed 67\% on any single language (see Table~\ref{tab:qclass-LU}). \texttt{Gemini-2.5} again leads (69.8 average), followed by \texttt{GPT-5} (52.5) \texttt{Claude-4.5} (49.8) and \texttt{DeepSeek-3.2} (46.3). Several medium-scale models punch above their parameter count: \texttt{Llama-4-Scout} (41.7, especially strong on Marathi) and \texttt{Gemma-3-27b} (42.2) occasionally outperform larger dense models on morphologically complex languages. Smaller models struggle severely, with many scores in the 20-35\% range; as evident in Table~\ref{tab:qclass-LU}. Sanskrit, Sanskrit+Eng, Dogri, Nepali and  Santali prove particularly challenging across the board.



A comparative analysis of GK and LU reveals distinct scaling behaviours. General knowledge accuracy correlates strongly with model scale (Pearson $r=0.91$ across the 20 evaluated models), whereas language-specific understanding shows a weaker but still substantial correlation ($r=0.73$). The correlation between the two dimensions is moderately high ($r=0.79$), indicating that top-tier factual reasoning usually co-occurs with stronger linguistic proficiency, yet meaningful divergences remain.

Several models achieve notably balanced performance. \texttt{Gemma-3-27b} (GK: 37.9, LU: 40.9) and \texttt{Mistral3.1-24b} (GK: 35.3, LU: 42.7) outperform many larger dense models on linguistic tasks, demonstrating that continued pre-training on diverse Indic corpora and instruction tuning can compensate for parameter count in morphosyntactically demanding settings. Similarly, MoE-based \texttt{Llama-4-Scout} reaches 41.8 on language understanding while scoring only 35.6 on general knowledge, highlighting the value of sparse architectures for grammar-heavy tasks.


These results underscore that future progress on low- and extremely low-resource Indic languages will require not only continued scaling but also (i) higher-quality and more diverse Indic pre-training data, (ii) dedicated cross-lingual continuation training, and (iii) architectural innovations that prioritise fine-grained morphological and syntactic representation.

\section{Conclusion}

We present \benchmark\ consisting of >13k questions for 11 low and extremely-low resource Indic languages. Our benchmark reveals that even the strongest contemporary LLMs, including frontier proprietary systems, remain far from reliable on graduate-level  questions in low- and extremely low-resource Indic languages, with no model surpassing 45\% average accuracy except Gemini-2.5. While larger closed models such as Gemini-2.5, GPT-5, Claude-4.5 and open-weight model, DeepSeek-3.2 consistently outperform open-weight alternatives. These findings indicate \benchmark\ as a challenging, human-curated evaluation suite that fills a key gap in existing Indic benchmarks and underline the need for  more balanced pretraining and adaptation for under-represented Indic languages.

\section{Limitations}
While \benchmark\ substantially broadens evaluation for low- and extremely low-resource Indic languages, the benchmark is derived exclusively from UGC-NET language papers in linguistics and literature, which biases the content toward academic exam styles and may not fully represent everyday or domain-general language use. Further, our evaluation is conducted using a log-likelihood-based multiple-choice setup in a zero-shot configuration, which may yield different results from alternative evaluation schemes such few-shot prompting, or chain-of-thought prompting, and therefore does not capture the full range of behaviors that these models might exhibit under richer prompting strategies. 

\section{Ethics}
All data described in this work was collected from publicly available resources.
\benchmark\ is released for non-commercial research use; code and annotation scripts are licensed under MIT. All annotation was conducted by consenting native speakers compensated at local rates, and no personally identifiable exam-candidate information appears in the benchmark. Users interested in commercial use should contact UGC-NTA for relevant permissions.

\bibliography{reference}
\appendix
\section{Annotation Setup}\label{sec:team}
For each language, we first collected official UGC-NET language papers and corresponding answer keys from the official website\footnote{https://ugcnet.nta.nic.in/} and converted all files into a uniform machine-readable format. Question papers appeared in double-column layouts, and many PDFs contained non-selectable text. Hence, all documents were processed through a proprietary OCR pipeline to extract text consistently across heterogeneous layouts and scan qualities. During this stage, each question was tagged with its language, subject, exam session, and year, and answer keys were aligned to their corresponding items.

Following OCR, human annotators manually corrected recognition errors, restored missing characters and diacritics, and normalised punctuation and spacing so that questions and options follow a pre-specified parsing format. Beyond surface correction, annotators standardised structural formatting—such as option markers, numbering schemes, and line breaks. Each final entry in \benchmark\ is stored with explicit fields for question text, question type, four answer options, correct option, language, subject, year, and unique question identifier.

Annotation was carried out by native speakers having expertise in the target language and having prior experience reading graduate-level exam material. Since many papers mix language-focused and knowledge-focused content, annotators were required to demonstrate strong proficiency in the relevant Indic script. All annotators received detailed written guidelines describing how to handle incomplete questions, ambiguous answer keys, and exam-specific conventions, as well as how to assign labels for language understanding (LU) versus knowledge-related (KR) items and for the six question types used in the benchmark.
Prior to full-scale annotation, annotators were given a sample dataset along with worked examples. They were then assigned trial files, which were reviewed and corrected by the reviewers. Feedback was provided iteratively, and only after demonstrating consistent accuracy were annotators assigned larger batches of data. 
For question–answer pairs, annotation quality followed a focused two-step review pipeline. In the first step, annotators corrected OCR errors in the question and options, aligned each item with the official answer key, and ensured that exactly one option was marked as correct. In the second step, a senior reviewer re-checked these corrected pairs against the original exam papers, verifying that the question text and the designated correct option was faithful to the source.
Annotators were compensated at a rate of \$1 per 10 questions.
\subsection{Quality Control and Verification}
\label{subsec: Quality Control and Verification}
A multi-stage quality control and verification pipeline was implemented to ensure that the dataset maintained integrity and linguistic precision. This first involved acquiring the primary source data directly from the official website of UGC NET, in high quality, natively generated PDFs, thereby ensuring high initial resolution and low extraction noise.

For the text extraction task, a approach was used based on script requirements. For languages using the Devanagari script, such as Sanskrit, Maithili, and Marathi, LLMWhisperer\footnote{\url{https://unstract.com/llmwhisperer/}} was used because it has been optimized for retaining complex Indic conjuncts. In contrast, non-Devanagari scripts, such as Gujarati and Odia, had significant extraction limits on multiple platforms. After comparing several options, Surya OCR \cite{paruchuri2025surya}  was implemented, but it had more script-specific mistakes that required intensive manual correction by subject area experts in Gujarati and Odia. All the Devanagari extractions then underwent parallel manual verification, wherein human annotators followed strict linguistic rules to correct ocr-induced errors and ensure structural consistency.

Quantification of final data quality was done via a post-verification audit on a randomized 5\% sample of the total file corpus. A sample of this size consists of 10,427 words and 61,547 characters, with 71 errors, which is a negligibly small WER of 0.68\% and CER of 0.12\%.

\section{Ethical considerations}
\label{sec:ethical-considerations}
This work introduces a new benchmark dataset derived from publicly available exam questions. As the dataset is built exclusively from existing public materials and involves no new data collection from human participants, no formal review by an Institutional Review Board (IRB) or equivalent ethics committee was required or conducted. Instead, the authors performed an internal assessment of data provenance, potential societal impacts, licensing considerations, and responsible use, in line with community standards for responsible machine learning research.
\subsection{Intended use and beneficiaries}
The benchmark is designed to support research in evaluating LLM understanding on Indic subjects. It aims to provide a challenging, realistic evaluation resource for the academic and AI research communities, as well as developers working on automated educational tools or knowledge-intensive systems. By releasing this dataset, we hope to foster progress in fair, transparent, and reproducible benchmarking of models at the graduate level, ultimately benefiting educators, students, and researchers seeking improved AI capabilities in knowledge assessment.
\subsection{Licensing and copyright considerations}
The original UGC-NET exam questions are publicly released by the National Testing Agency (NTA), Government of India, primarily for candidate preparation and result-related purposes (available on \href{ugcnet.nta.ac.in}{ugcnet.nta.ac.in}). No explicit redistribution license (e.g., Creative Commons) or statement prohibiting non-commercial research use is provided on the official website.\\

We consider our work as a  transformative use (curated selection, cleaning, and reformatting into a machine-readable benchmark for 13,207 questions focused on Indian languages) permissible under fair dealing provisions of the Indian Copyright Act, 1957, for non-commercial research and educational purposes. We do not redistribute raw original question papers/PDFs; only processed question-answer pairs with metadata are released. The derived IndicParam dataset is made available under the Creative Commons Attribution 4.0 International (CC-BY-4.0) license, requiring attribution to both this work and the original UGC-NET source.

\begin{table*}[!]
\centering
\small
\begin{tabular}{|p{\dimexpr\textwidth-2\tabcolsep-2\arrayrulewidth\relax}|}
\hline
\multicolumn{1}{|c|}{\textbf{Zero-Shot Prompt}} \\
\hline

\begin{minipage}[t]{\linewidth}
\ttfamily
Prompt = f"""Question: \{[`question\_text']\}\\

Options:\\
A) \{[`option\_a']\}\\
B) \{[`option\_b']\}\\
C) \{[`option\_c']\}\\
D) \{[`option\_d']\}\\

The above question is written in \{language\} language. Please analyze the question and options carefully, and select the correct answer. Respond ONLY with one letter (A, B, C, or D) corresponding to the correct option. Do not provide any explanation or additional text.""" 

\end{minipage}
\\
\hline
\end{tabular}
\caption{Zero-shot prompt applied across all models for evaluation.}
\label{tab:zero-shot-prompting}
\end{table*}
\begin{table*}[!]
\centering
\resizebox{.75\textwidth}{!}{%
\begin{tabular}{@{}cccccccc@{}}
\toprule

\textbf{Language} & \textbf{MCQ}  & \textbf{A\&R} & \textbf{List Matching} & \textbf{Blank Filling} & \textbf{IS}  & \textbf{Ordering} & \textbf{Total} \\ \midrule
Nepali            & 957           & 1             & 38                     & 5                      & 25           & 12                & 1038           \\
Marathi           & 933           & 83            & 139                    & 1                      & 54           & 35                & 1245           \\
Gujarati          & 345           & 172           & 208                    & 0                      & 128          & 191               & 1044           \\
Odia              & 379           & 40            & 70                     & 1                      & 22           & 65                & 577            \\ \midrule
Maithili          & 1265          & 0             & 0                      & 21                     & 0            & 0                 & 1286           \\
Konkani           & 1069          & 30            & 134                    & 19                     & 45           & 31                & 1328           \\
Santali           & 772           & 36            & 46                     & 11                     & 2            & 6                 & 873            \\
Bodo              & 481           & 288           & 270                    & 17                     & 0            & 257               & 1313           \\
Dogri             & 223           & 153           & 217                    & 1                      & 231          & 202               & 1027           \\
Rajasthani        & 1114          & 4             & 11                     & 14                     & 13           & 34                & 1190           \\
Sanskrit          & 1210          & 1             & 39                     & 34                     & 11           & 20                & 1315           \\
Sans-Eng  & 905           & 3             & 13                     & 33                     & 14           & 3                 & 971            \\ \midrule
\textbf{Total}    & \textbf{9653} & \textbf{811}  & \textbf{1185}          & \textbf{157}           & \textbf{545} & \textbf{856}      & \textbf{13207} \\ \bottomrule
\end{tabular}
}
\caption{Different question types across all languages in \benchmark. A\&R refers to Assertion and Reason type question, IS refers to identification of incorrect statement. `Sans-Eng' denotes a separate set of Sanskrit-English code-mixed question-answer pairs which forms a separate set.} 
\label{tab-app:stats-qtypes}
\end{table*}


\begin{table*}[!]
\centering
\resizebox{0.75\textwidth}{!}{
\begin{tabular}{@{}clcccccc@{}}
\toprule
\textbf{Size}                & \textbf{Model} & \textbf{MCQ}        & \textbf{List Matching} & \textbf{A\&R}       & \textbf{Blanks}     & \textbf{IS}         & \textbf{Ordering}   \\ \midrule
\multirow{4}{*}{Small-size}  & Qwen2.5-3B     & 30.3                & 32.6                   & 25.6                & 26.1                & 27.7                & 30.5                \\
                             & Llama3.2-3B   & 26.9                & 25.2                   & 29.3                & 28.7                & 22.4                & 31.2                \\
                             & Gemma3-4b     & 28.7                & 27.6                   & {\ul \textit{30.8}} & 28                  & 27.5                & 28.7                \\
                             & Qwen3-4B       & {\ul \textit{31.5}} & {\ul \textit{27.8}}    & 27.1                & {\ul \textit{31.8}} & {\ul 28.4}          & {\ul \textit{32.6}} \\ \midrule
\multirow{4}{*}{Medium-size} & Aya-8b        & 29.1                & 28.9                   & 32.3                & 34.4                & 25.5                & 27.9                \\
                             & Llama3.1-8B   & 30.5                & 23.2                   & 32.7                & 35                  & 24.4                & 27.6                \\
                             & Gemma3-12b    & 35.1                & {\ul \textit{35.9}}    & 36.7                & 33.1                & {\ul \textit{37.1}} & 34.8                \\
                             & Mistral3.1-24b & 35.7                & 34.3                   & 27.1                & 33.8                & 35.8                & {\ul \textit{37}}   \\
                             & Gemma3-27b    & {\ul \textit{38.1}} & 32.8                   & {\ul \textit{37.2}} & {\ul \textit{38.9}} & 36.3                & 35.7                \\ \midrule
\multirow{4}{*}{Large-size}  & Aya-32b        & 33.8                & 36.4                   & 31.4                & {\ul \textit{36.9}} & 27.2                & 31.3                \\
                             & Qwen3-32B      & 34.6                & 24.6                   & 32.8                & 29.9                & 31.9                & 29.8                \\
                             & Llama3.3-70B  & {\ul \textit{37}}   & {\ul \textit{39.7}}    & {\ul \textit{41.6}} & 34.4                & {\ul \textit{40.7}} & {\ul 38.1}          \\ 
                             & DeepSeek-3.2       & 43.2                & \textbf{47.2}          & \textbf{44.5}       & {\ul 45.9}          & 38                  & {\ul \textit{37.1}} \\
                             \midrule
\multirow{3}{*}{MoE}         & Qwen3-A3B      & {\ul \textit{39.9}} & 33.1                   & 37.9                & {\ul \textit{40.1}} & {\ul \textit{35.6}} & 34                  \\
                             & Llama-4-Scout  & 37.5                & \textit{28.3}          & 34.6                & 35                  & 30.8                & 26.4                \\
                             & gpt-oss-120b   & 37.9                & {\ul \textit{38}}      & {\ul \textit{38.7}} & 38.2                & 32.5                & {\ul \textit{37.4}} \\ \midrule
\multirow{4}{*}{Closed}      
                             & GPT-5          & {\ul 46.1}       & {41.3}             & {41.4}          & {\ul 47.8}       & {43.3}          & {\ul 41.6}       \\
                             & Grok4-fast    & 40.4                & 39.2                   & 33.7                & 36.9                & 38.5                & 36.9                \\
                             & Claude-4.5     & {43.9}          & 39                     & 40                  & 37.6                & {\ul 45.7}       & 36.6                \\ 
& Gemini-2.5 & \textbf{\textit {57.2}} & \textbf{\textit {65.6}} & \textbf{\textit {55 }} &  \textbf{\textit {59.5}} & \textbf{\textit {58.5}} & \textbf{\textit {51.4}}  \\                            
                             \bottomrule
\end{tabular}
}
\caption{Question-type wise performance of all models on \benchmark.}
\label{tab:qtype-perf}
\end{table*}
\section{Question Types-wise results}\label{app:sec-qtype}

Table \ref{tab-app:stats-qtypes} presents the distribution of question types across languages, and Table \ref{tab:qtype-perf} provides the corresponding performance results for these question types.

\subsection{Question Types}
Table \ref{tab-app:stats-qtypes} provides a detailed breakdown of the types of questions present in each language. MCQs dominate the distribution, but a substantial number of list matching, Assertion–Reasoning(A\&R), identification of incorrect statements (IS), ordering and blank filling questions contribute to different question types.
\subsection{Question type-wise results}
Table \ref{tab:qtype-perf} shows the performance of each model on \benchmark\ by question type.  In the smaller-sized models, \texttt{Qwen3-4B} demonstrates better performance in a majority formats of questions, which includes MCQ, IS, fill in the blanks and Ordering. This model outperforms alternative 3–4B models, including \texttt{Llama3.2-3B} and \texttt{Gemma3-4B}.  In the medium-size model category, \texttt{Gemma3-27B} shows the best overall performance, with considerable improvements in MCQ, A\&R, and fill in the blanks. At the same time, \texttt{Gemma3-12B} and \texttt{Mistral3.1-24B} also show competitive results in List Matching and Ordering, respectively.  Large models show a considerable improvement in their ability to handle different question types. \texttt{Llama3.3-70B} model performs best on most question types, especially MCQ, list matching, A\&R and IS. \texttt{DeepSeek-3.2} performs best on MCQ,  list matching, A\&R shows good results compared to other large-sized models in the category. Within MoE models, \texttt{Qwen3-A3B} performs best on MCQ, IS and fill in the blanks type questions. On the other hand, \texttt{gpt-oss-120b} shows the best results on list matching, A\&R, and ordering related questions.  Closed-source models consistently outperform their open-source counterparts. \texttt{GPT-5}, achieves best scores across almost all question categories, including MCQ, fill in the blanks, list matching, A\&R and ordering. In contrast, \texttt{Claude-4.5} demonstrates best performance for IS questions.  \texttt{Grok4-fast} performance is  consistently lower than that of \texttt{GPT-5} and \texttt{Claude-4.5}.

\section{Zero shot prompting template}
In Table \ref{tab:zero-shot-prompting}, we provide the zero-shot prompt used while evaluating all the models. 

\section{Performance on GK and LU questions }
The trends observed in Table \ref{tab:qclass-gk}  and Table \ref{tab:qclass-LU}  show that accuracy improves consistently with increasing model scale, with frontier models achieving the strongest results. As discussed in Section \ref{sec:gk-performance}, general knowledge tasks display steady but moderate gains, whereas Section \ref{sec:lang-performance} highlights far greater variability in language-specific tasks, reflecting the difficulty of deep multilingual grounding. Architectural improvements, including MoE routing, enable mid-sized models narrow the performance gap with larger systems.






\begin{table*}[]
\resizebox{\textwidth}{!}{
\begin{tabular}{@{}lccccccccccc@{}}
\toprule
\textbf{Model} & \textbf{Gujarati}      & \textbf{Konkani}       & \textbf{Maithili}      & \textbf{Marathi}       & \textbf{Oriya}         & \textbf{Rajasthani}    & \textbf{Sanskrit}      & \textbf{Sans-Eng}      & \textbf{Dogri}         & \textbf{Nepali}        & \textbf{Santali}    \\ \midrule
Qwen2.5-3B     & {\ul \textit{29.3}}    & {\ul \textit{31.5}}    & 29.2                   & 28.2                   & 29.1                   & 27.1                   & 31.7                   & {\ul \textit{35.5}}    & 32.2                   & 27.3                   & 33.5                \\
Llama3.2-3B   & 26.2                   & 28.2                   & 21.1                   & 28.3                   & 27.4                   & 26.3                   & 27.8                   & 27.9                   & 28                     & 25.1                   & 27.3                \\
Gemma-3-4b     & 26.3                   & 29.1                   & 23.8                   & 29.2                   & 30.4                   & 25.3                   & 29.5                   & 32.3                   & 30.4                   & 28.2                   & 31.5                \\
Qwen3-4B       & 27.3                   & 29.3                   & {\ul \textit{31.5}}    & {\ul \textit{29.7}}    & {\ul \textit{32.6}}    & {\ul \textit{27.4}}    & {\ul \textit{34.5}}    & 33.6                   & {\ul \textit{35.6}}    & {\ul \textit{28.9}}    & {\ul \textit{33.9}} \\ \midrule
Aya--8b        & 27.6                   & 26.6                   & 29                     & 28.3                   & 27.8                   & 31.5                   & 29.7                   & 30.7                   & 31.2                   & 27.2                   & 30.1                \\
Llama-3.1-8B   & 27.3                   & 29.1                   & 27.9                   & 31.8                   & 28.9                   & 29.3                   & 32.4                   & 34.2                   & 25.9                   & 24.6                   & 30.4                \\
Gemma-3-12b    & {\ul \textit{33.9}}    & 35.5                   & 29.7                   & 37.5                   & 33                     & 32.6                   & 39                     & 43.1                   & 36.7                   & 34.8                   & 33.5                \\
Mistral3.1-24b & 32.3                   & 31.7                   & {\ul \textit{36.3}}    & 32.1                   & 28                     & 32.3                   & {\ul \textit{41.3}}    & 45.2                   & 38                     & 34.8                   & 35.8                \\
Gemma-3-27b    & 31.3                   & {\ul \textit{36}}      & 33.2                   & {\ul \textit{38.9}}    & {\ul \textit{37.8}}    & {\ul \textit{36.9}}    & 39.9                   & {\ul \textit{49.4}}    & {\ul \textit{37.8}}    & {\ul \textit{39.4}}    & {\ul \textit{36.7}} \\ \midrule
Aya-32b        & 28.4                   & 32.5                   & 31.7                   & 32.5                   & 28.3                   & 33                     & 36.3                   & 40.6                   & 35.3                   & 32.5                   & 35.5                \\
Qwen3-32B      & 35.9                   & 34.2                   & 31.5                   & 40                     & 38.5                   & 29.5                   & 40.1                   & 49.5                   & 43.1                   & 33.6                   & 39.7                \\
Llama-3.3-70B  & 33.6                   & 37.2                   & 35.2                   & 42.7                   & {\ul \textit{39.3}}    & 36.6                   & 42.8                   & 52.1                   & 37.5                   & 35.8                   & 39                  \\
DeepSeek-3.2   & {\ul \textit{39.6}}             & {\ul \textit{37.3}}    & {\ul \textit{39.7}}    & {\ul 43.9}             & 38.7                   & {\ul \textit{37.4}}    & {\ul \textit{53.5}}             & {\ul \textit{62.9}}             & {\ul \textit{42.8}}    & {\ul \textit{39.5}}    & {\ul 40.7}       \\ \midrule
Qwen3-A3B      & 30.5                   & 28.3                   & 29.6                   & 34.9                   & 33.9                   & 29.1                   & 38.2                   & 44.7                   & \textit{35.8}          & 30.8                   & 35.3                \\
Llama-4-Scout  & 27.4                   & 34.4                   & 30                     & {\ul \textit{42.1}}    & 32.2                   & 32.2                   & {\ul \textit{44.3}}    & 49.4                   & 30                     & 34                     & 35.7                \\
gpt-oss-120b   & {\ul \textit{37}}      & {\ul \textit{34.9}}    & {\ul \textit{33.9}}    & 38.7                   & {\ul \textit{36.3}}    & {\ul \textit{35.5}}    & 41.5                   & {\ul \textit{50.2}}    & {\ul \textit{38.7}}    & {\ul \textit{36.1}}    & {\ul \textit{35.7}} \\ \midrule
GPT-5          & {\ul 40.1} & {38.5}             & {\ul 41.4} & {\ul 47.7} & {\ul 44.1} & {\ul 40.8} & {\ul 57.8} & {\ul 66.3} & {\ul 45.1} & {41.5}             & { 40.1}          \\
Grok-4-fast    & 35.9                   & 37.1                   & 36.5                   & 38.8                   & 38.3                   & 37.1                   & 45.2                   & 55.9                   & 39.9                   & 40.6                   & 38.1                \\
Claude-4.5     & 37                     & {\ul 41.9} & {40.3}             & 42.9                   & 39.8                   & 39.7                   & 48.8                   & 59.5                   & {44.9}             & {\ul 43.4} & {40.3}       \\

Gemini-2.5 & \textbf{\textit {51.8}} & \textbf{\textit {47.6}} & \textbf{\textit {50.4}} & \textbf{\textit {54.3}} & \textbf{\textit {60.9}} & \textbf{\textit {64}} & \textbf{\textit {74.4}} & \textbf{\textit {80.3}} & \textbf{\textit {52.4}} & \textbf{\textit {50.4}} & \textbf{\textit {52.3}} \\

\bottomrule
\end{tabular}
}
\caption{Performance of models on knowledge related question category.}
\label{tab:qclass-gk}
\end{table*}

\begin{table*}[]
\resizebox{\textwidth}{!}{
\begin{tabular}{@{}lccccccccccc@{}}
\toprule
\textbf{Model} & \multicolumn{1}{c}{\textbf{Gujarati}} & \multicolumn{1}{c}{\textbf{Konkani}} & \multicolumn{1}{c}{\textbf{Maithili}} & \multicolumn{1}{c}{\textbf{Marathi}} & \multicolumn{1}{c}{\textbf{Oriya}} & \multicolumn{1}{c}{\textbf{Rajasthani}} & \multicolumn{1}{c}{\textbf{Sanskrit}} & \multicolumn{1}{c}{\textbf{Sans-Eng}} & \multicolumn{1}{c}{\textbf{Dogri}} & \multicolumn{1}{c}{\textbf{Nepali}} & \multicolumn{1}{c}{\textbf{Santali}} \\ \midrule
Qwen2.5-3B     & {\ul 66.7}                            & 27.3                                 & 25.4                                  & 20.7                                 & 27.4                               & 32.4                                    & {\ul \textit{32.6}}                   & {\ul \textit{29.7}}                   & {\ul \textit{31.9}}                & 26.2                                & 30.3                                 \\
Llama3.2-3B   & 16.7                                  & {\ul \textit{51.5}}                  & 28.5                                  & 34.5                                 & 25.6                               & 26.4                                    & 28.8                                  & 27.9                                  & 27.7                               & 30.8                                & 28.3                                 \\
Gemma-3-4b     & 16.7                                  & 30.3                                 & 30.8                                  & 43.1                                 & 35.9                               & 33.6                                    & 27.7                                  & 31.5                                  & 30.9                               & {\ul \textit{32.3}}                 & 33.3                                 \\
Qwen3-4B       & 50                                    & 39.4                                 & {\ul \textit{33.1}}                   & {\ul \textit{46.6}}                  & {\ul \textit{42.7}}                & {\ul \textit{36.7}}                     & 30.7                                  & 33.3                                  & 29.8                               & 27.7                                & {\ul \textit{33.3}}                  \\ \midrule
Aya--8b        & 50                                    & 39.4                                 & 38.5                                  & 27.6                                 & 29.1                               & 29.1                                    & 28.4                                  & 23.4                                  & 33                                 & 32.8                                & {\ul \textit{36.4}}                  \\
Llama-3.1-8B   & 16.7                                  & 45.5                                 & 39.2                                  & 39.7                                 & 30.8                               & 38.2                                    & 31.4                                  & 32.4                                  & 28.7                               & 29.7                                & 32.3                                 \\
Gemma-3-12b    & 16.7                                  & 42.4                                 & {\ul \textit{49.2}}                   & 56.9                                 & 40.2                               & 41.5                                    & 29.5                                  & 31.5                                  & {\ul \textit{39.4}}                & 34.9                                & 35.4                                 \\
Mistral3.1-24b & \textbf{\textit{83.3}}                         & 45.5                                 & 46.9                                  & 48.3                                 & 25.6                               & 41.2                                    & {\ul \textit{34.8}}                   & 35.1                                  & 37.8                               & 35.4                                & 35.4                                 \\
Gemma-3-27b    & 0                                     & {\ul \textit{57.6}}                  & 44.6                                  & {\ul \textit{60.3}}                  & {\ul \textit{47.9}}                & {\ul \textit{47.6}}                     & 33.3                                  & {\ul \textit{46.8}}                   & 36.7                               & {\ul \textit{39}}                   & 36.4                                 \\ \midrule
Aya-32b        & {\ul \textit{33.3}}                   & {\ul 57.6}                           & 46.2                                  & 46.6                                 & 31.6                               & 38.8                                    & 31.1                                  & 27                                    & 34.6                               & 35.4                                & 36.4                                 \\
Qwen3-32B      & 33.3                                  & 45.5                                 & 56.2                                  & 56.9                                 & 49.6                               & 44.8                                    & 36                                    & 41.4                                  & 44.7                               & 41                                  & 40.4                                 \\
Llama-3.3-70B  & 16.7                                  & 42.4                                 & {\ul \textit{57.7}}                   & 58.6                                 & 44.4                               & 42.4                                    & 37.5                                  & 47.7                                  & 36.2                               & {\ul \textit{43.1}}                 & \textit{\textbf{48.5}}               \\
DeepSeek-3.2   & \textit{16.7}                         & {\ul \textit{48.5}}                  & 56.9                                  & {\ul 58.6}                           & {\ul \textit{55.6}}                & {\ul \textit{47}}                       & {\ul 42.8}                            & {\ul 49.5}                            & {\ul 45.7}                         & 40.5                                & {\ul 47.5}                           \\ \midrule
Qwen3-A3B      & {\ul \textit{50}}                     & 30.3                                 & 43.1                                  & 48.3                                 & 36.8                               & 38.2                                    & 30.7                                  & 40.5                                  & \textit{35.6}                      & 34.9                                & 40.4                                 \\
Llama-4-Scout  & 16.7                                  & 45.5                                 & 42.3                                  & \textit{\textbf{70.7}}               & 46.2                               & 41.5                                    & {\ul \textit{42.8}}                   & {\ul \textit{43.2}}                   & 32.4                               & 36.9                                & {\ul \textit{41.4}}                  \\
gpt-oss-120b   & 33.3                                  & {\ul \textit{54.5}}                  & {\ul \textit{50.8}}                   & 60.3                                 & {\ul \textit{53.8}}                & {\ul \textit{49.7}}                     & 34.8                                  & 31.5                                  & {\ul \textit{44.1}}                & {\ul \textit{41.5}}                 & 35.4                                 \\ \midrule
GPT-5          & {\ul \textit{33.3}}                   & {\ul 60.6}               & {\ul 60.8}                & {\ul \textit{63.8}}                  & {\ul 66.7}             & {\ul 54.5}                              & {\ul 43.2}                            & {\ul 51.4}                & {\ul 48.4}             & {\ul 51.8}              & 43.4                                 \\
Grok-4-fast    & 33.3                                  & 51.5                                 & 52.3                                  & 60.3                                 & 50.4                               & 44.2                                    & 33.7                                  & 40.5                                  & 41.5                               & 42.6                                & {\ul \textit{44.4}}                  \\
Claude-4.5     & 33.3                                  & 51.5                                 & {\ul 60.8}                            & 63.8                                 & {\ul 60.7}                         & {\ul 55.8}                  & {\ul 43.6}                & 46.8                                  & 44.1                               & {\ul 45.6}                          & 41.4                                 \\

 Gemini-2.5 & \textbf{\textit {33.3}} & \textbf{\textit {78.8}} & \textbf{\textit {85.4}} & \textbf{\textit {84.5}} & \textbf{\textit {78.6}} & \textbf{\textit {73.3}} & \textbf{\textit {64.4}} & \textbf{\textit {80.2}} & \textbf{\textit {62.2}} & \textbf{\textit {59}} &\textbf{\textit {57.6}}     \\ 
\bottomrule
\end{tabular}
}
\caption{Performance of models on language understanding (LU) question category.}
\label{tab:qclass-LU}
\end{table*}

\section{Examples questions in \textit{\benchmark}}
Table \ref{tab:examplequestions} presents examples of six distinct types of questions for each language used in \benchmark. 


\begin{table*}[t]
\centering
\footnotesize
\setlength{\tabcolsep}{1.5pt}
\renewcommand{\arraystretch}{1.05}

\begin{tabular}{
p{5.2cm}
p{1.9cm}
p{1.9cm}
p{1.9cm}
p{1.9cm}
>{\centering\arraybackslash}p{0.5cm}
>{\centering\arraybackslash}p{1.1cm}
>{\centering\arraybackslash}p{0.6cm}
>{\centering\arraybackslash}p{0.8cm}
}
\toprule
\textbf{Question} &
\textbf{option(a)} &
\textbf{option(b)} &
\textbf{option(c)} &
\textbf{option(d)} &
\textbf{Ans} &
\textbf{Type} &
\textbf{Class} &
\textbf{Lang} \\
\midrule

\writehi{1.बालिवधस्य वर्णनमस्ति रामायणस्य} &
\writehi{सुन्दरकाण्डे} &
\writehi{किष्किन्धा काण्डे} &
\writehi{अरण्यकाण्डे} &
\writehi{बालकाण्डे} &
b & MCQ & G & san \\
\midrule

\writehi{2.'मन्त्रिपरिषदं द्वादशामात्यान्कुर्वीत' इति कस्य मान्यता ?} &
\writehi{बार्हस्पत्यानाम्} &
\writehi{कौटिल्यस्य} &
\writehi{औशनसाम्} &
\writehi{मानवानाम्} &
d & MCQ & G & san \\
\midrule

\writehi{3.अधोऽडिकतानां समीचीनमुत्तरं चिनुत -}\newline
(a) \writehi{ सरमा-पणि 1. बृहदारण्यकोप-सम्वाद: निषत्}\newline
(b) \writehi{स्वाध्यायान्मा 2. ऋग्वेदस्य}
\writehi{प्रमदः दशममण्डले} \newline
(c)\writehi{ कल्प: 3. तैत्तिरीयोपनिषत्}\newline
(d)\writehi{ आत्मनस्तु 4. हस्त:}\newline
\writehi{कामाय सर्वं प्रियं भवति}
\writehi{(a) (b) (c) (d)} &
1 3 2 4 &
4 2 3 1 &
2 3 4 1 &
3 2 1 4 &
c & Order & G & san \\
\midrule

\writehi{4.भाशा दी सभनें थमां लौहकी इकाई ऐ} : &
\writehi{ध्वनि} &
\writehi{ध्वनिग्राम} &
\writehi{रूपग्राम} &
\writehi{वाक्य} &
a & MCQ & L & doi \\
\midrule

\writehi{5.पैहूली चंदी च दित्ती गे दियें प्रविष्टियों दा दूई चंदी दियें प्रविष्टियें कन्नै स्हेई मिलान करो} :\newline
\writehi{चंदी-1} \quad \writehi{चंदी-2}\newline
\writehi{(अ) डोगरी काव्य-चर्चा} (i) \writehi{प्रो. रामनाथ शास्त्री}\newline
\writehi{(ब) परख-पड़ताल} (ii) \writehi{शिवनाथ}\newline
\writehi{(स) डोगरी साहित्य दा इतिहास} (iii) \writehi{प्रो. चम्पा शर्मा}\newline
\writehi{(द) बाबा जित्तो} (iv) \writehi{ओम गोस्वामी}\newline
\writehi{कोड : (अ) (ब) (स) (द)} &
(ii) (i) (iii) (iv) &
(iv) (ii) (i) (iii) &
(iii) (iv) (ii) (i) &
(i) (iii) (iv) (ii) &
c & LM & G & doi \\
\midrule

\writehi{6.इक लड़ाई होर' ते 'बंजर' दे रचेता न} : &
\writehi{मदन मोहन शर्मा ते मोहन सिंह ।} &
\writehi{मोहन सिंह ते नरसिंह देव जम्वाल ।} &
\writehi{मोहन सिंह ते मदन मोहन शर्मा ।} &
\writehi{जितेन्द्र शर्मा ते दीनू भाई पंत ।} &
c & Order & G & doi \\
\midrule

7. \begin{gujarati} બન્ને યાદીની વિગતો સરખાવી સાચાં જોડકાં બનાવો \end{gujarati}:\newline
(a) \begin{gujarati}શેષાદ્રિ\end{gujarati} \quad (i) \begin{gujarati}વિજયરાય વૈદ્ય \end{gujarati}\newline
(b) \begin{gujarati}રામ વૃંદાવની\end{gujarati} \quad (ii) \begin{gujarati}ત્રિભુવનદાસ લુહાર\end{gujarati}\newline
(c) \begin{gujarati}ત્રિશુળ\end{gujarati} \quad (iii) \begin{gujarati}રાજેન્દ્ર શાહ\end{gujarati}\newline
(d) \begin{gujarati}મયૂરાનંદ\end{gujarati} \quad (iv) \begin{gujarati}ખબરદાર\end{gujarati}\newline
(a) (b) (c) (d) &
(i) (iv) (iii) (ii) &
(iv) (iii) (ii) (i) &
(ii) (iv) (iii) (i) &
(iii) (ii) (i) (iv) &
b & LM & G & guj \\
\midrule

8. \begin{gujarati}નીચેનાં વિધાનોને કાર્યકારણસંબંધે તપાસો\end{gujarati} :\newline
(A) \begin{gujarati}'વદતોવ્યાઘાત' મધ્યમપદલોપી સમાસ છે.\end{gujarati}\newline
(R) \begin{gujarati}મધ્યમપદલોપી સમાસ સર્વપદપ્રધાન છે. \end{gujarati} &
(A) \begin{gujarati}અને \end{gujarati}(R) \begin{gujarati}બન્ને સાચાં છે.\end{gujarati} &
(A) \begin{gujarati}અને \end{gujarati}(R) \begin{gujarati}બન્ને ખોટાં છે. \end{gujarati} &
(A) \begin{gujarati}સાચું છે અને \end{gujarati}(R) \begin{gujarati}ખોટું છે. \end{gujarati} &
(A) \begin{gujarati}ખોટું છે અને \end{gujarati}(R)\begin{gujarati} સાચૂં છે. \end{gujarati} &
b & A\&R & G & guj \\
\midrule

9.\begin{gujarati}નીચેનામાંથી સુસંગત વિગતજૂથ જણાવો \end{gujarati}: &
\begin{gujarati}વીસમી સદી, વૈશ્વાનર, નટમંડળ \end{gujarati} &
\begin{gujarati}ગુજરાત વિદ્યાસભા, ગુજરાતી સાહિત્ય પરિષદ, ગુજરાત સાહિત્ય અકાદમી\end{gujarati} &
\begin{gujarati}ઊડણ ચરકલડી, સોયનું નાકું, મનીષા\end{gujarati} &
\begin{gujarati}ખીચડી, અંત:સ્ત્રોતા, આવતીકાલનો સૂરજ \end{gujarati} &
b & MCQ & G & guj \\

\bottomrule
\end{tabular}
\end{table*}

\begin{table*}[t]
\centering
\footnotesize
\setlength{\tabcolsep}{1.5pt}
\renewcommand{\arraystretch}{1.05}

\begin{tabular}{
p{5.2cm}
p{1.9cm}
p{1.9cm}
p{1.9cm}
p{1.9cm}
>{\centering\arraybackslash}p{0.5cm}
>{\centering\arraybackslash}p{1.1cm}
>{\centering\arraybackslash}p{0.6cm}
>{\centering\arraybackslash}p{0.8cm}
}
\toprule
\multicolumn{9}{c}{\textit{Table~\ref{tab:examplequestions} continued}} \\
\midrule
\textbf{Question} &
\textbf{option(a)} &
\textbf{option(b)} &
\textbf{option(c)} &
\textbf{option(d)} &
\textbf{Ans} &
\textbf{Type} &
\textbf{Class} &
\textbf{Lang} \\
\midrule

\writehi{10.फावो तो पर्याय वेंचून काडून वाक्य पूर्ण करात.}\newline
\writehi{सोळाव्या शेंकड्यातल्या रामायणाच्या हातबरपांत} &
\writehi{ ठांयीं ठांयीं पुर्तुगेज उतरां मेळटात.} &
\writehi{कांयच पुर्तुगेज उतरां मेळनात.} &
\writehi{संस्कृत उतरां भरसून पुर्तुगेजीक कोंकणीची} \writehi{सया मारता.} &
\writehi{देशी आनी अपभ्रंशी भारतीय भासो मेळटात.} &
b & Blank filling & G & gom \\
\midrule

\writehi{11.भाशीक नादांचो भाशेचे बांदावळीचे नदरेंतल्यान अभ्यास करपी व्याकरणाक कितें म्हणतात ?} &
\writehi{ अर्थविज्ञान} &
\writehi{नादविज्ञान} &
\writehi{वाक्य-विचार} &
\writehi{स्वनीम-विचार} &
d & MCQ & G & gom \\
\midrule

\writehi{12. काळोकिट' हो समास हांतलो खंयचो ?} &
\writehi{उपमावाचक कर्मधारय समास} &
\writehi{विशेशण उभयपद समास} &
\writehi{द्वंद्व समास} &
\writehi{बहुव्रिहि समास} &
b & MCQ & G & gom \\
\midrule

\writehi{13. हुनकासँ भेंट भेल छल' लिखने छथि} &
\writehi{मणिपद्म} &
\writehi{अमर} &
\writehi{सुमन} &
\writehi{मधुप} &
a & MCQ & G & mai \\
\midrule

\writehi{14. "शिव की ..... विष पचाय यदि लितथि नहि माथे ।"} &
\writehi{देखितथि} &
\writehi{बुझितथि} &
\writehi{करितथि} &
\writehi{सकितथि} &
d & Blank filling & G & mai \\
\midrule

\writehi{15. एकावली परिणय' में अंगीरस अछि} &
\writehi{शान्त} &
\writehi{अद्भुत} &
\writehi{ हास्य} &
\writehi{श्रृंगार} &
d & MCQ & L & mai \\
\midrule

\writehi{16. पुढीलपैकी कोणते नाटक विजय तेंडुलकर यांनी लिहिले नाही ?} &
\writehi{गृहस्थ} &
\writehi{सूर्यास्त} &
\writehi{बेबी} &
\writehi{श्रीमंत} &
b & MCQ & G & mar \\
\midrule

\writehi{17. पुढीलपैकी कोणती कादंबरी ऐतिहासिक नाही ?} &
\writehi{वज्राघात} &
\writehi{कालिकामूर्ति} &
\writehi{ सावळ्या तांडेल} &
\writehi{दुर्दैवी रंगू} &
b & IS & G & mar \\
\midrule

\writehi{18. पुढील साहित्यकृतींचा कालानुसार क्रम लावा.} &
\writehi{धग, किडलेली माणसे, औदुंबर, स्मृतिचित्रे,} &
\writehi{किडलेली माणसे, औदुंबर, स्मृतिचित्रे, धग,} &
\writehi{धग, औदुंबर, स्मृतिचित्रे, किडलेली माणसे,} &
\writehi{औदंबर, स्मृतिचित्रे, किडलेली माणसे, धग,} &
d & Order & G & mar \\

\bottomrule
\end{tabular}
\end{table*}

\begin{table*}[t]
\centering
\footnotesize
\setlength{\tabcolsep}{1.5pt}
\renewcommand{\arraystretch}{1.05}

\begin{tabular}{
p{5.2cm}
p{1.9cm}
p{1.9cm}
p{1.9cm}
p{1.9cm}
>{\centering\arraybackslash}p{0.5cm}
>{\centering\arraybackslash}p{1.1cm}
>{\centering\arraybackslash}p{0.6cm}
>{\centering\arraybackslash}p{0.8cm}
}
\toprule
\multicolumn{9}{c}{\textit{Table~\ref{tab:examplequestions} continued}} \\
\midrule
\textbf{Question} &
\textbf{option(a)} &
\textbf{option(b)} &
\textbf{option(c)} &
\textbf{option(d)} &
\textbf{Ans} &
\textbf{Type} &
\textbf{Class} &
\textbf{Lang} \\
\midrule

\writehi{19. शास्त्रीय मार्क्सवादी दृष्टिमा आइडियोलजी भन्नाले के बुझिन्छ ?} &
\writehi{भ्रमपूर्ण चेतना ।} &
\writehi{सैद्धान्तिक आस्था} &
\writehi{प्रचलित विचारधारा ।} &
\writehi{पुस्तकहरू पढेर पाइने ज्ञान ।} &
a & MCQ & G & nep \\
\midrule

\writehi{20. वाक्यमा शब्दहरू माझ रहने सम्बन्धलाई के भनिच्छ ?} &
\writehi{ विन्यासक्रमी सम्बन्ध ।} &
\writehi{सहचारक्रमी सम्बन्ध ।} &
\writehi{व्यतिरेकी सम्बन्ध ।} &
\writehi{परस्परव्यापि सम्बन्ध ।} &
a & MCQ & L & nep \\
\midrule

\writehi{21. नेपाली भाषामा प्रयोग हुने 'चोलो', 'पटुका', 'मुन्धुम', 'बाउसे' शब्दहरू कुन भाषा परिवारबाट आएका हुन् ?} &
\writehi{ भारोपेली ।} &
\writehi{ भोट-बर्मेली ।} &
\writehi{आग्नेय ।} &
\writehi{द्रविड़ ।} &
b & MCQ & L & nep \\
\midrule

\writehi{22. "राजस्थानी भाषा रौ विगसाव नागर अपभ्रंश सूं हुयौ है।" ओ कथन है -} &
\writehi{डॉ. एल.पी. टेस्सिटोरी} &
\writehi{ डॉ. ग्रियर्सन} &
\writehi{ डॉ. सुनीतिकुमार चटर्जी} &
\writehi{डॉ. सुकुमार सेन} &
b & MCQ & G & Raj \\
\midrule

\writehi{23. राजस्थानी अनुवाद री दीठ सूं किसौ अनुवाद रौ जोड़ौ गलत है ?} &
\writehi{गांधीजी री आत्मकथा - आईदानसिंह भाटी} &
\writehi{भरथरी शतक - मनोहर शर्मा} &
\writehi{मेघदूत - नारायणसिंह भाटी} &
\writehi{रचाव - चेतन स्वामी} &
d & IS & G & Raj \\
\midrule

\writehi{24. बुगचौ' सबद रौ सही अर्थ है :} &
\writehi{लुगाइयां रै कपड़ा राखण रौ कपड़े रौ थेलौ} &
\writehi{लुगाइयां री लकड़ी री पेटी} &
\writehi{लुगाइयां री कपड़ा राखण री लोह री पेटी} &
\writehi{मड़दां रै कपड़ा राखण री लकड़ी री पेटी} &
a & MCQ & L & Raj \\
\midrule

\writehi{25. स हि कदाचिद् वाच्ये विधिरूपे प्रतिषेधरूपः' अत्र स इत्यनेन कोऽभिप्रेतः ?} &
\writehi{वाच्यार्थः} &
\writehi{लक्ष्यार्थः} &
\writehi{व्यङ्गार्थ:} &
\writehi{ तात्पर्यार्थः} &
c & MCQ & L & sans-mix \\
\midrule

\writehi{26. कौषीतकि उपनिषद् कस्य वेदस्य अस्ति ?}\newline
\writehi{कौषीतकि उपनिषद् किस वेद की है ?}\newline
Of which Veda is \writehi{कौषीतकि उपनिषद् ?} &
\writehi{अथर्ववेदस्य} &
\writehi{ सामवेदस्य} &
\writehi{ ऋग्वेदस्य} &
\writehi{ कृष्णयजुर्वे दस्य} &
c & MCQ & G & sans-mix \\
\midrule

\writehi{27. अधोलिखितेषु केन सह कस्य सम्बन्धः ? उचितां तालिकां चिनुत -}\newline
(a) \writehi{सत्कार्यवाद:} (i) \writehi{न्यायवैशेषिकाणाम्}\newline
(b) \writehi{परमाणुवादः (ii) बौद्धानाम्}\newline
(c) \writehi{विवर्तवादः} (iii) \writehi{संख्यानाम्}\newline
(d) \writehi{विज्ञानवाद:} (iv) \writehi{अद्वैतवेदान्तिनाम्}\newline
(a) (b) (c) (d) &
(iii) (ii) (i) (iv) &
(iii) (iv) (i) (ii) &
(iii) (i) (iv) (ii) &
(iii) (i) (ii) (iv) &
c & LM & G & sans-mix \\

\bottomrule
\end{tabular}
\end{table*}

\begin{table*}[t]
\centering
\footnotesize
\setlength{\tabcolsep}{1.5pt}
\renewcommand{\arraystretch}{1.05}

\begin{tabular}{
p{5.2cm}
p{1.9cm}
p{1.9cm}
p{1.9cm}
p{1.9cm}
>{\centering\arraybackslash}p{0.5cm}
>{\centering\arraybackslash}p{1.1cm}
>{\centering\arraybackslash}p{0.6cm}
>{\centering\arraybackslash}p{0.8cm}
}
\toprule
\multicolumn{9}{c}{\textit{Table~\ref{tab:examplequestions} continued}} \\
\midrule
\textbf{Question} &
\textbf{option(a)} &
\textbf{option(b)} &
\textbf{option(c)} &
\textbf{option(d)} &
\textbf{Ans} &
\textbf{Type} &
\textbf{Class} &
\textbf{Lang} \\
\midrule

\writehi{28. सेदाय हापड़ामको काथा लेकाते होड़ होपोनकोवाक् पारिस दो ओको जायगा रे को हाटिज लेदा ?} &
\writehi{चाम्पा} &
\writehi{चाय} &
\writehi{सासाडबेडा} &
\writehi{ जारपी} &
c & MCQ & G & sat \\
\midrule

\writehi{29. लातार रे ओल आकान थापित काथा (A) आर ओजे (R) रेनाक् ओका गादेल बाड़ती सुहिया ?}\newline
\writehi{थापित (A) : भारतीय नोजोर ते गायान साँवहेतु दो सुक रे मुच़ात लाकतीयाना ।}\newline
\writehi{ओजे (R) : चेदाक् जे नोवा काथा हों पछिम दिसोम रेन साँपहेतुहिया नोजोर ते हों नापाया ।} &
\writehi{ (A) आर (R) बाङसुहीया} &
\writehi{ (A) सुहीया, (R) बाडसुहीया} &
\writehi{ (A) बाडसुहीया, (R) सुहीया} &
\writehi{(A) आर (R) बानार सुहीया} &
d & A\&R & G & sat \\
\midrule

\writehi{30. ओका पारसी घारोंचू रेनाक् ओका पारसी काना ? ओना सुही मिलाव में ।}\newline
\writehi{सूची - I} \quad \writehi{सूची - II}\newline
a.\writehi{आर्य भाषा परिवार} \quad 1. \writehi{तिब्बती}\newline
b. \writehi{अग्नेय भाषा परिवार} \quad 2. \writehi{ माल्तो}\newline
c. \writehi{ द्राविड़ भाषा परिवार} \quad 3.\writehi{ मुण्डा}\newline
d. \writehi{ चीनी भाषा परिवार} \quad 4. \writehi{बंगला} \newline
\writehi{कूट} : a b c d &
1 2 3 4 &
4 3 2 1 &
3 1 4 2 &
2 4 3 1 &
b & LM & G & sat \\
\midrule

31. \begin{odia}ମୁକୃ ଅକ୍ଷର ଶେଷରେ କଣ ଥାଏ \end{odia}? &
\begin{odia}ଅକ୍ଷର \end{odia} &
\begin{odia} ସୂରଧୂନି \end{odia} &
\begin{odia}ବ୍ୟଞ୍ଜନଧୂନି \end{odia} &
\begin{odia}ସୁର୍କିଧୂନି \end{odia} &
b & MCQ & L & ory \\
\midrule

32. \begin{odia}ପ୍ରକାଶକାଳ ଅନୁଯାୟୀ ଉପଯୁକ୍ତ କ୍ରମ ନିର୍ଦ୍ଧାରଣ କର ।\end{odia} &
\begin{odia}ଉଷା, ଚିଲିକା, ପାର୍ବତୀ, ମହାଯାତ୍ରା\end{odia} &
\begin{odia}ଚିଲିକା, ଉଷା, ପାର୍ବତୀ, ମହାଯାତ୍ରା\end{odia} &
\begin{odia}ପାର୍ବତୀ, ଚିଲିକା, ଉଷା, ମହାଯାତ୍ରା\end{odia} &
\begin{odia}ଉଷା, ପାର୍ବତୀ, ଚିଲିକା, ମହାଯାତ୍।\end{odia} &
d & Order & G & ory \\
\midrule

33. \begin{odia}କେଉଁଟି ବ୍ୟୋମକେଶ ତ୍ରିପାଠୀଙ୍କ ନାଟକ ନୁହେଁ \end{odia}? &
\begin{odia}ସିଂହଦ୍ୱାର\end{odia} &
\begin{odia}ଜାଗରଣ \end{odia} &
\begin{odia}ମୁଠାଏମାଟି\end{odia} &
\begin{odia}ସୁନାଫରୁଆ\end{odia} &
c & IS & G & ory \\
\midrule

\writehi{34. बाहायनाने गाहायाव होनाय मेथाइफोरनि बाहायजानाय सम फारि साजायनायखौ सायख -}\newline
I. \writehi{बैसागो मेथाइ}\newline
II. \writehi{दोहोरोम सिबिनाय मेथाइ}\newline
III. \writehi{बेलाड}\newline
IV. \writehi{मैंगं खानायाव खननाय मेथाइ} &
II III IV I &
III II I IV &
IV III I II &
I II IV III &
b & Order & G & brx \\
\midrule

\writehi{35. अलंबार' लाइसिखौ सोर सुजुदोंमोन} &
\writehi{ मदाराम ब्रह्म} &
\writehi{सुकुमार बसुमतारी} &
\writehi{सतिस चन्द्र बसुमतारी} &
\writehi{प्रम'द चन्द्र ब्रह्म} &
d & MCQ & G & brx \\
\midrule

\writehi{36. गाहायनि बुंफोरनाय (A) आरो जाहोन (R) खौ गेबें ना गोरोन्थि बाहायनानै सायख :}\newline
\writehi{बुंफोरनाय (A) : आथिखालाव मोननाय बासिराम जोहोलाव बेलाडखौ थारै आबुं बेलाड एबा natural ballad बुंथावा ।}\newline
\writehi{ जाहोन (R) : मानोना बे बेलाडा लिरनाय महराव फोसावजानाय ।} &
\writehi{(A)-आरों (R) मोननैबो गेबें} &
\writehi{(A)-आ गेबें आरो (R)-आ गोरोन्थि} &
\writehi{(A) आरो (R) मोननैबो गोरोन्थि} &
\writehi{(A)-आ गोरोन्थि आरो (R) आ गेबें} &
b & A\&R & G & brx \\

\bottomrule
\end{tabular}

\caption{Sample questions from all languages in \benchmark\ along with their
options, answers, question types, and question classes, with languages
indicated using ISO-639 codes. We use \textit{Order} to denote ordering-type
questions, and the abbreviations \textit{G} and \textit{L} refer to General
Knowledge (GK) and Language Understanding (LU) questions respectively.}
\label{tab:examplequestions}
\end{table*}
\end{document}